# Children's Acquisition of Tail-recursion Sequences: A Review of Locative Recursion and Possessive Recursion as Examples

Xiaoyi Wang[b], Chenxi Fu[c], Yang Caimei[a]* and Zhuang Ziman[d]

[a]*School of Foreign Language, Soochow university, Suzhou, China;* [b]*School of Foreign Language, Soochow university, Suzhou, China;* [c]*School of Foreign Language, Soochow university, Suzhou, China;* [d] *Ulink High School of Suzhou Industrial Park, Suzhou, China*

Corresponding author: Yang Caimei   Email: cmyang@suda.edu.cn

Xiaoyi Wang[b] and Chenxi Fu[c] are co-first authors due to the same contribution to this review.

# Children's Acquisition of Tail-recursion Sequences: A Review of Locative Recursion and Possessive Recursion as Examples

Recursion is the nature of human natural language. Since Chomsky proposed generative grammar, many scholars have studied recursion either theoretically or empirically. However, by observing children's acquisition of tail recursion sequences, we can verify the nativism of language supported by universal grammar and reveal the cognitive mechanism of human brain. To date, our understanding of children's acquisition path of recursion and influencing factors still remain controversial. This systematic review summarizes the research of tail recursive sequence by taking possessive recursion and locative recursion as examples, focusing on the experimental methods, acquisition paths, and influencing factors of tail recursive sequence. The current behavioural experiments reveal that, the debate about children's performance revolves around: 1) Gradual acquisition or synchronous acquisition. 2) symmetry or asymmetry between the acquisition of locative recursion sequences and possessive recursion sequences. We presume that children can acquire recursion quickly in a short period of time thanks to the language acquisition device, though there are also scholars who believe that a third factor also plays a role.

Keywords: tail-recursion sequence; possessive recursion; locative recursion; child language acquisition

Subject classification codes: include these here if the journal requires them

## 1. Introduction

Recursion, a central idea in mathematics and computer science, refers to the grammar of human natural language as a finite set of recursive rules that can generate an infinite number of linguistic expressions. Chomsky believes that recursion is the essence of language, which can distinguish human language from animal forms of communication. (Yang, 2020) Over the years, the importance of recursion has been widely studied both theoretically and empirically, including in children's language acquisition. We believe that by observing the performance of children's acquisition recursion process, the mechanism and process of children's language development can be better analysed from the deep level, such as recursive understanding, cognitive load difficulties and so on. Taking possessive recursion and locative recursion under tail recursion classification as examples, this paper will sort out the theories related to recursive acquisition and analyse the empirical research methods and research results. To find out the acquisition path and other outcomes in children's acquisition of tail-recursion sequences, this overview paper comprehensively reviews the cutting-edge research regarding children's acquisition of the possessive recursion (Elmo's sister's ball) and locative recursion (the bird on the crocodile in the water). We mainly focus on the age development path and limiting factors of recursion in children's acquisition.

Section 2 introduces theories that have been put forward or proved in research about children's acquisition of possessive or locative recursion. Section 3 reviews research contents and section 4 addresses research methods used in previous research. In section 5, the acquisition path and constraints will be analysed. Display quotations of over 40 words, or as needed.

## 2. Theories

Before taking a further step into theories about children's acquisition, we shall review the previous definition of tail recursion. Tail recursion/ end recursion (including left and right-branching recursion), which refers to a type of recursive structure where the recursive call appears at the end of the sentence, along with centre-embedded recursion (CER) are called two main types of recursion occurring in natural language. (Fedor A, 2011) Rohrmeier et al. (2014) also discuss the term "tail-recursion" and it seems that, in their opinion, "tail-recursion" is conjunction/coordination since they only list two expressions The green, red, yellow, purple … balls are in the box and This sentence continues and continues and continues as the illustrations of "tail-recursion" in natural languages. (Yang C.M., in press) In Yang's (in press) research, tail-recursion rules are a subtype of recursive rules that can produce different models of sequences that recursively expand to the tail side or sides of the sequences. "α →< α  β >" is a left branch rule and "α →< α  β >" is a right branch rule. Tail recursion sequences are formed using these rules. For instance, in English structures of recursive possession and recursive locatives, all belong to tail-recursion sequences: the former can be either X's Y's Z (e.g. John's sister's ball) or X of Y of Z (e.g. the ball of the sister of John) and the latter is mostly right-handed recursion (e.g., The bird on the alligator in the water). This paper follows Yang's recursive system, that is, both the possessive recursion sequences and locative recursion sequences belong to tail-recursion sequences. Next, we will systematically outline theories proved by children's acquisition of tail-recursion sequences.

*2.1 Indirect& direct recursion*

Roeper and Oseki argue that there is an acquisition path from direct recursion (i.e. Coordination) to indirect recursion (i.e. embedding). Roeper's (2011) proposals for a two-step acquisition path pointed out that before children acquire adult-like indirect recursion (iterative embedding of one phrasal category inside another of the same type), they prefer to make direct recursion (i.e. conjunction or simple merge) at the age around 5. (Usha Lakshmanan, 2022) This two-step path, in Roeper's opinion, is a way of reconciling the fundamental role of recursion in language with children's attested difficulties with different forms of embedding.

Since the inception of this theory, there have been numerous empirical studies conducted around it. Sevcenco et al. (2015) predicted that conjunction will appear first for every structure at a young age, and then re-appear when the structure becomes complex. Their research found that children below the age of 6 prefer to give conjunctive syntactic answers than indirect recursive structure. 5-,6- and 7-year-old children give more indirect recursion readings for 2-level embeddings than for 3-level embeddings.

Larisa Avram et al. (2021) then investigated Hungarian-Romanian bilingual children and their results for 2L1 and L1 Romanian support Roeper's (2011) view according to which direct recursion is the acquisition default.

*2.2 Three-step acquisition path*

Nakato and Roeper (2021) proposed a three-level path in first language acquisition of recursion: 1. Search for overt morphological elements 2. Search for overt lexical elements 3. The ability to interpret recursive structures even without the presence of identical overt elements. (Hossein Rouhalamini Najafabadi et al., 2023) This hypothesis can be proved by Sevcenco's research, in which for 5yr-old children, overtly marked recursive configurations (relative clauses + that) got more accuracy than covertly marked ones (modification by means of 7-year-olds at earlier ages. Hossein Rouhalamini Najafabadi et al.'s research also proved this by testing Persian recursion because Persian is a language that overtly marks recursive locatives with a functional element. Mao et al.'s (2024) research showed that in Mandarin, Explicit functional centre De (的) promotes children's understanding of recursive structure: children aged 4;1 successfully mastered two-level DeP recursive structure with an explicit De, while a level 2 DeP recursive structure with an implicit De can only be acquired until the child is 4;3.

*2.3 Recursion-only hypothesis (ROH)*

The recursion-only hypothesis is a hypothesis concerning the unique nature of human language. ROH argues that in a narrow sense, language function includes recursion as a unique computational mechanism and recursion is a unique capacity of human language and an innate characteristic of the human brain, which means that recursion doesn't exist in other animal's communication system. (Yang, 2014) Shi et al. (2019) investigated children's acquisition of possessive structure "NP1 De(的) NP2 De(的) NP3". In their research, children could acquire a 2-level possessive recursion structure at an early stage (4 years old), which is similar to Mao et al.'s research (2024). The outcome is corroborated by the three features stated by Crain (1991) and Crain et al. (2017): early emergence, universality, and language independence, proving the ROH thereby.

*2.4 The maturation theory*

The maturation theory is based on Chomsky's Universal Grammar, which proposed that similar to other physiological organs, children's language capacity needs certain developmental stages for maturation. (Yang, 2014) Specifically, the Maturation Theory argues that some language phenomena are not comprehended by learning, but the gradual maturation. For instance, the Optional Infinitive Stage reflects parameters that have not yet been fully set during specific stages of neurological development. Children usually exhibit different grammatical capacities in different developmental stages, which is closely related to the maturation of specific brain regions. In the following part of this article, you will see different ways of children's language acquisition and performance, which also confirms this hypothesis.

**3. Research contents**

As shown in Figure 1, the current empirical research on possessive recursion and locative recursion in academia involves 12 languages with 32 research in total, in which English and Japanese research account for the vast majority and other minor languages such as Wapichana, Tamil and Persian are also measured. These structures are mainly first to fourth-level tail-recursion sequences.

English-based research mainly studies two-level recursion. Gentile (2003) began to study the acquisition of left-branching possessive recursion (Cookie Monster's sister's picture) in children. Limbach and Adone (2010) investigated the acquisition of two-level recursive possession (Jane's father's bike) in English-speaking children. Pérez-Leroux et al. (2012) recruited monolingual English-speaking children between the ages of 3 and 5 and observed whether the children produced phrases with two-level recursive possession like [[[N's] N's] N]] through a small task of looking at a picture and listening to a story. Sevcenco et al. (2015) classified direct recursion (conjunction) and indirect recursion (recursion), as well as explicit and implicit structures, to explore whether children are more likely to make juxtaposed interpretations or prefer explicit structures in locative recursion. In the study of the Giblin et al. (2019) experiment participants were 30 4-year-old monolingual English-speaking children, and the study also used an elicited truth-value judgment task to guide subjects' output of level 2 genitives (e.g., 海盗的青蛙的饼干/Pirate's Frog's Cookie).

Japanese-based research delves into both possessive recursion and locative recursion. The Japanese language has a marker "の" similar to the Chinese "的" and the Englis "'s" to indicate possessive relationships. It can serve as both a genitive marker and a locative connective, with category ambiguity which leads to the uniqueness of acquisition. Currently, Japanese-related studies involve second-, third-, and fourth-level recursion. For instance, Fujimori (2010) and Terunuma et al. (2017) used の-containing 2-4 level recursion structures to ask questions like "ミカの弟の靴は何色ですか" (What color are Mika's brother's shoes)? and "働物園の車の上の猫は何色ですか" (What color is a cat on the car in the zoo). Inada et al. (2015) find that Japanese-spoken children are better at locative recursion than possessive recursion though both structures can be a challenge. The experiment by Nakajima et al. (2014) provided evidence that the accuracy of possessive recursion reached merely 30.6% compared with that of locative recursion's accuracy of about 58.3%. However, Pérez-Leroux et al. (2023) made a different conclusion that for 5 and 6-year-old children as well as adults, the acquisition of possessive recursion is higher than locative recursion while the level of recursion structure is the same.

To date, the academic community has also shown interest in languages other than English. Relevant studies in Mandarin tail-recursion sequences revolve around 2 level possessive recursion. Yang (2014) first explored 3- to 10-year-old children's acquisition and production of nested recursion with picture-matching elicitation task. Her team then did a series of studies in possessive recursion with different semantic types, recursion levels, and research methods. Another research team that has done great studies in possessive recursion is Zhou Peng's team: Shi et al. (2019) recruited 30 monolingual Mandarin children under 4 years old to investigate the production of 2-level possessive recursion with functional head "的" (de). In Iain Giblin's (2019) second study, the Mandarin children's acquisition of 2-level possessive recursion (海盗的青蛙的饼干/Pirate's Frog's Cookie) was researched by watching the picture while hearing the story and then making choices. Apart from these, Li et al. (2020) did a comprehensive experimental analysis of Mandarin children's acquisition of 2- and 3-level possessive recursion structure. In their study, 30 children aged 3-6 were divided into two groups according to their age. The experiment includes two phases. In the maturation phase, researchers described the possessive relationship in the picture with recursion structure to children. Then in the test phase, children were elicited to produce corresponding possessive relationships, for instance, "机器人的蛇的狮子的饼干/ Robot's snake's lion's biscuit."

The acquisition of recursive possessive structure has also attracted attention in other languages, Roberge et al. (2018) drew on the material design of Pérez-Leroux et al. (2012) and changed the original English material to French (Le ballon de la sœur d' Elmo. /

Elmo's sister's ball) for testing. Grohe et al. (2021) explored the acquisition of German recursive possessive structures (e.g. Marias Nachbars Buch/Maria's Neighbour's Book). Research has also looked at recursions that are acquired early by children in niche languages such as Wapishana (Xa'apauran Cedrick dadukuu minhayda'y yza bala-n tan?/What colour is the ball of Cedrick's sister's friend?), but children only reach adult performance after the age of seven (Leandro et al., 2014); Usha Lakshmanan (2023) explored Tamil children's understanding of recursive possessive structures from one to four levels (AshavooDu tambiooDu pirenDooDu naayooDu balloon enna niram?). What color is Asha's brother's friend's dog's ballon?) Toth, A. (2017) and Hossein Rouhalamini Najafabadi et al. (2023) tested second-order position recursion in Hungarian native children and second-order, third-order, and fourth-order position recursion in Persian native children by performing tasks (rearranging animals), respectively. Bleotu A. C. (2021) uses picture-matching tasks to expose Romanian-speaking children to both recursive and coordination structures. In addition, there is a small amount of literature on the recursive acquisition of bilingual children. Experiments by Pérez-Leroux et al. (2017) show that for bilingual children (Spanish-English), the distribution of response types is similar between bilingual and monolingual children under recursive conditions, except for the frequency of incomplete responses. Larisa Avram et al. (2020) conducted a comparative study on recursive embedding of second-level.

    Comparatively speaking, at this stage, the research is still mainly focused on English, and there are relatively few studies on other languages. There is a lack of further analyses of the structure of possessive structures in each language, and fewer studies take into account the influence of semantics or the structure of the language itself on the acquisition, and some of the analyses on the disturbing factors are only on the level of theoretical inference, without the validation of empirical studies, so we must study recursive

possessive acquisition by children more comprehensively. If we want to study children's acquisition of recursive possession more comprehensively or to further prove the recursion-only hypothesis, there is still plenty of room for research on other languages.

## 4. Research design

Since most of these empirical studies are behavioural experiments on children, and it is not possible to accurately capture the activity in children's brains in behavioural experiments, and children's cognition can only be responded to through children's responses, ensuring that each response corresponds to a unique interpretation behind it is a very important point to ensure the rigor of the design of behavioural experiments. In the behavioural experiments on recursive possession and recursive locatives, scholars over the years have designed a variety of experiments to ensure the control of experimental conditions, the standardization of the operation process, and the accuracy of data collection, to obtain more reliable and valid experimental results.

In the field of child language acquisition, there are two primary experimental methods: production experiments and comprehension experiments. One of the production techniques is the referential elicitation task, which is also a usual method to examine the acquisition of locative recursion as well as possessive recursion. (Crain, S., & Thornton, R., 1998) This type of design mainly contains pictures and prompts. During the experiment, researchers will first introduce a series of pictures and the relationship between each referential, and then elicit participants to describe or recognize these pictures with a target response. Through this, the language structure as well as the language acquisition process can be better observed to mark down important phenomena. (Crain, S., & Thornton, R., 1998）In the process of testing the acquisition of locative recursion, to elicit a complex NP containing three related nouns, Pérez-Leroux (2012) designed a simple referential task. In the task, children hear a brief description of the picture, accompanied by an image, and finally a question (e.g. What is broken and flat?) The target response is a two-level recursive possessive structure (e.g. Elmo's sister's ball (is flat)), and if the description is not given using a two-level recursive possessive structure, the response may be ambiguous. This experimental design was later adapted and translated to test the acquisition of spatial relations in Japanese, Spanish, and bilingual children. On this basis, the coding systems are designed based on the task content and expectation from the perspective of syntax and semantics to detect the influence on acquisition, which can be an innovation point. For instance, the syntactic coding system includes (a)single NP, Level 1, 2 level 1 and level 2, and the semantic coding system are composed of five conditions which are "INCOMPLETE", "SEQUENTIAL", "ALTERNATIVE", "NON-EMBEDDING" and "TARGET". (Peterson et al., 2015) (Pérez-Leroux & Pettibone, 2017) ( Pérez-Leroux et.al, 2023). Roberge (2018) adopted the referential elicitation task used by Pérez-Leroux et al. (2012) and adapted the English material to French. In sum, subjects can directly understand tasks via picture presentation as well as a brief introduction to referential possessive or locative relationship so that children can produce recursive structure accordingly. For instance, in Pérez-Leroux et al.'s (2023) research, each trial was organized with one picture and even a short story, which was designed to improve subjects' interests considering children's age. However, such an experimental design faces a problem: the whole process takes a long time up to 30 minutes due to the story segment.

  The truth-value judgment task is a comprehension technique used to study children's understanding of the meaning of test materials. It is divided into two categories: the yes/no judgment and the reward/punishment judgment. In recent years, truth-value judgment task and their variants have also been favoured by many scholars (Iain Giblin 2019). Initially, Crain and McKee (1985) applied the truth-value judgment task to investigate the meanings that children can and cannot attribute to sentences. (Crain, S., & Thornton, R.,1998) Based on their method, Shi (2019) modified this design into an "elicited production task with a truth value judgment component": The experiment involved two experimenters, one of whom told a story to a child with a toy and the other of whom played a blindfolded animal with a doll. In the task, subjects and small animals listen to a story, and the experimenter asks the animal a question after the story is over. The task of the child is to tell the animal whether its answer is correct or not. After children's yes/no judgment of the animal puppet's answer, they need to take it a step further. If correct, they need to repeat the animal's words; if not, children need to tell where's wrong and help them correct. This innovation based on the TVJ experimental paradigm not only increases the interest in the experiment process, but also creates contextual conditions for the rational use of recursion for children, reduces the difficulties and errors in children's understanding and output, and better demonstrates the acquisition of children's recursive structure. Some scholars have also made innovations in design. Manami (2020) and Li Daoxin et al. (2022) allowed children to add a familiarity phase to the experiment. In the familiarity stage, subjects are required to repeat the basic recursive structure. This repetition can not only avoid misunderstanding of experimental materials in the experiment, which may lead to ambiguous interpretation, such as parallel interpretation, but also will not promote children's output ability, which will promote the further detection of whether children can produce the possessive recursive structure in the following experimental stage.

In addition to the Truth-value judgment task, the act-out task is also a kind of comprehension experiment. This paradigm regards children's comprehension as the acquisition of recursion structure, usually with relatively little work. The first experiment on recursive possessives was designed by Gentile (2003) in which each child was given two pictures about characters in Sesame Street. While researchers ask questions like "Can you show me Cookie Monster's sister's picture?", Children will watch related pictures and make choices. Since then, there has been a lot of research acquiring such a design. In Limbach and Adone's research (2010), children and adult English learners were asked to make comprehension and judgment of recursion structure such as "Jane's father's bike" after hearing a story and picking the correct one among many similar pictures. Likely, Fujimori（2010）shows the subjects the pictures and then asks questions after telling them the story. His questions based on the content of the picture using a recursive possessive structure containing levels two to four, for example, "ミカのお兄さんの友達の犬のボールは何色ですか/What color is Mika's brother's friend's dog ball ?" An experiment paradigm like this is relatively simple due to less difficulty. Leandro and Amaral (2014) adapted the corpus of Roeper (2011) into Wapicchana and English, and Lakshmanan (2023) designed an experiment adapted from Roeper (2011) and Terunuma et al (2017). The children were shown pictures of two families and told the story in Tamil, and the subjects were asked to assess their understanding of the first to fourth-level possessive (e.g. AshavooDu tambiooDu pirenDooDu naayooDu balloon enna niram? / What color is Asha's brother's friend's dog balloon? Lakshmanan (2023) and Bleotu (2021) urge children to choose between a picture corresponding to a recursive interpretation and a picture corresponding to a coordinated interpretation after hearing a sentence containing a recursive or juxtaposed structure. In such experimental paradigms, determining whether a child has acquired a recursive possessive structure is highly dependent on the subject and picture design, so ensuring rigor is particularly important. Roeper et al. (2012) used the same method to test locative recursion in Japanese. Since then, many experimental designs (Terunuma 2012, 2018) have been adapted from the design of Roeper (2012). An example is the point-out task in investigating Tamil children's acquisition of locative recursion in which participants need to point out or say the item's name after hearing the story and prompts "Show me the crocodile in the water in the pond". (Lakshmanan, 2022) In addition, to investigate the position recursive acquisition of Hungarian and Romanian bilingual children, Larisa Avram (2020) adopted a performance task, that asked children to arrange animal pictures into different arrangements according to the prompts, which improved their attention to a certain extent and aroused their participation and enthusiasm. The experiment was adapted from Sevcenco et al.'s experiment in 2015, in which participants were asked to rearrange the animals and repeat the prompts they heard. However, in some experiments involving a long story background and complex character relationships, subjects' answers will be affected by the memory load, and it may not be enough for children to establish such abstract representation of possessive relationships only through static pictures and short verbal descriptions of possessive relationships in graphs, especially those involving multiple layers of possessive relationships in recursive structures (Shi et al., 2019).

Despite of research design above mentioned, there are some new research methods in possessive and locative research. Based on the biological characteristics of the human body, these experimental designs try to find more direct evidence of acquisition recursion according to the sound signals and brain waves emitted, so they have higher accuracy and authority. Manami's research (2020) is interested in whether Japanese children exhibit adult-like prosody in complex noun phrases, whether they use prosody to mark phrase attachment, and whether there is a correlation between prosodic ability and the ability to produce recursive noun phrases. The research method includes examining children's prosodic performance at the lexical level to determine if they have the prerequisites for further analysis; analyzing the pitch peak (f0 peak) decline pattern in recursive structures, especially the pitch differences in AAA and UAU recursive structures; and testing pitch reset or prosodic enhancement in recursive and non-recursive structures. In Marcus Maia et al.'s research (2021), ERP (Event-Related Potentials) was used to investigate the neurophysiological underpinnings of processing differences between PP (Prepositional Phrase) embedding and coordination. The ERP tests aimed to detect online processing effects by recording brain activity while participants processed different types of syntactic constructions.

**5. Research Outcomes**

Figure 2 comprehensively exhibits children's developmental stage of acquisition of tail-recursion sequences. Most of the participants are children who are between the ages of three to six. According to the percentage shown in the figure, we could conclude that apparently, acquisition of tail-recursion sequences exists in various languages, although with desynchrony of acquisition stage among different languages and asymmetry between possessive recursion sequences and locative recursion sequences. Besides, for bilingual children, their acquisition of tail-recursion sequences enjoys certain latency compared with monolingual children.

*5.1 Gradual acquisition & synchronous acquisition*

Terunuma (2017) considered that children would understand two-level, three-level, and four-level recursion in an adult-like manner one by one, that is, in a step-by-step development. However, Fujimori (2010) has shown that Japanese-speaking children can acquire two- and three-level recursion at once; Terunuma and Nakato (2013) observed that Japanese-speaking children can interpret 3-level recursion and 4-level recursion at the same time. Pérez-Leoux (2012) found that producing coordinated nominals is easy for children but recursive complex nominals are difficult. The same phenomenon is also observed in Sevcenco et al. (2015) and Mao et al. (2024). Pérez-Leoux et al. reckons that coordination lacks s-selectional requirements compared to recursion or doesn't need semantic integration. They also mentioned that perhaps coordination does not involve merging but some (possibly) more primitive form of concatenation or aggregation.

For the gradual and synchronous acquisition phenomenon, Terunuma and Nakato (2020) gave two analyses. The first analysis is that, for a step-by-step developmental path, they think children don't know how to properly arrange and nest multiple possessive phrases in a syntactic structure. In this analysis, only a single possessive phrase is possible in the form of a noun modifier, taking possessive recursion as an example. In the second stage, children can produce both 1-POSS structure and non-recursive 2-POSS structure. Then finally in the third stage, the DP substitution within POSSPs triggers recursive structures in which a DP contains another DP through a DP- possessive, which means children could truly produce a 2-level possessive recursion. The second analysis argues that a correlation is expected between the acquisition of recursive possessives and that of multiple Negative Polarity Items or WH- operators. Specifically, there is a licensing mechanism allowing them to handle various complex structures, including recursive possessives, without needing a gradual transition once they acquired.

*5.2 Asymmetry between possessive recursion sequences& locative recursion sequences*

Inada (2015) analyzed and compared the research outcome of Nakajima et al. (2014), Terunuma and Nakato-Miyashita (2013) and Hollebranse and Roeper (2014) and drew the conclusion that embedded Recursion structure is unavailable uniformly to English and Japanese children, but Japanese children show a relatively high accuracy in interpreting recursive locative expressions. Inada contributes this to the morphological and syntactical difference triggered by the morpheme 'の'(NO), which allows the non-unique semantic explanation. For instance, when '-の' is used as a possessive marker, it indicates a possessive relationship. For example, "東京の親戚" (Tokyo-no shinseki) means "a relative from Tokyo." On the other hand, when "-の" is used as a locative marker, it indicates a locative relationship. For example, "公園の中のベンチの上の猫" (kooen-no naka-no benchi-no ue-no neko) means "the cat on the bench in the park." The function of "の"(NO) was also discussed in Terunuma (2017), in which semantic variation of no's existence is confirmed while regarding spatial relations in locative recursion. However, the 4-year-old children's performance contradicts this prediction.

　　Pérez-Leoux et al. (2024) argue that a purely structural explanation is not available. Their study found that although different kinds of recursive structures differ grammatically, the experimental results showed that these structures did not differ significantly in children's performance, but both admitted and children showed similar difficulties in dealing with locative and part-whole relations. It can be reasoned that the challenge of recursion is not just at syntax level, but the design of the third factor. Children's productivity of recursion depends on the Level-1 embedding (or its statistical signature), which is a trigger for acquisition of recursion of that configuration. In detail, Children firstly need to comprehend the specific relation and the associated marker, and then figure out which form allows for the embedding or iteration, for example, Elmo's sister's ball. Another example is their research about German children's DP acquisition in 2020. The findings indicate that although the German possessive -s is widely used, children are not inclined to overapply it in recursive structures. Therefore, we can suppose that a certain amount of input is required for acquisition. Apart from this, previous studies (e.g., Arslan et al. 2017 and Pérez-Leroux et al. 2018b) have shown that sentence repetition ability, as a measure of linguistic memory, is significantly correlated with the ability to generate recursive structures. However, there is no direct evidence proving that working memory decides their acquisition of recursion structure.

In addition, Terunuma (2017) assumed that the number of choices in the picture may affect children's performance due to children need to do forced choice among a large number of items, especially for 4-level recursion.

## 6. Discussion

In this paper, the main research results in the field of locative recursion and possessive recursion in the past 20 years are reviewed. At present, children's acquisition of tail recursion shows the characteristics of gradualism and asymmetry. Progressive refers to the results of most papers showing that children's acquisition of recursion is positively correlated with recursion level and age. Asymmetry refers to the fact that most papers show that the acquisition between locative recursion and possessive recursion is not synchronized ---- children who have learned possessive recursion do not necessarily learn locative recursion. Whether it is influenced by grammatical structure, or by the third factor including cognitive load, memory processing and behavioural testing methods, there is no single positive answer in the academic community. In addition, since almost all existing studies have adopted different methods of behavioural experiments, there are too many factors that affect the rigor of behavioural experiments. As shown in the figure 2, the number of interference items, experiment duration and children's attention, and some behavioural experiments indirectly regard understanding as acquisition, which may also lead to different results. We then compare the 4 and 5 age groups in the same language, and take the second-level possessive recursion structure as an example. In English, it is obvious that Compared with the referential elicitation method (Pérez-Leoux et al., 2012 &2024), the truth value judgment method (TVJ) has a higher accuracy rate (Iain Giblin et al., 2019). We attribute this to the fact that the adoption of puppets and dialogue, which stimulate children's interest. In addition, A second-level possessive recursion structure with incorrect semantics is set, which is more acceptable to children than full self-production:

Question to puppet: Which hotdog got knocked over?

Blindfolded puppet: I can't see but let me guess. Gecko's hotdog got knocked over!

Target response: No, Gecko's koalas' hotdog got knocked over!

(Iain Giblin et al., 2019)

Now we turn to Pérez-Leroux's research in 2012. Before the experiment, children were not told to answer with a specific possessive recursion structure. If Iain Giblin et al.'s experiment presents certain syntactic cues in the last sentence at the end of the puppet dialogue, then in this experiment the children are completely autonomous in producing the target structure according to the problem：

Genitive task

Here is Elmo. This is his sister. And here is Bart and that's his sister. They each have a ball. Their sisters are carrying their balls too. They are all going together to the basketball court. But look! Oh, Oh.

Prompt: What is broken and flat?

Target: Elmo's sister's ball (is flat)

(Pérez-Leroux, 2012)

After constructing a specific situation, the child heard a "what" question, but the child did not know what specific structural response the experimenter was expecting, so it was only necessary for them to let the researcher know their answer in a clear way after hearing the question. So, we can see that in addition to being able to clearly identify

unlearned conjunction and 1-level possessive modification, there are also the following answers：

Non-embedding strategies

a. Um, his little sister's, a broken ball (RSMB 3;08) (Possessive Pronoun)

b. this one, look, flowers (ARA 4;09) (Appositional NPs)

c. She looks like, um, the dog has the hat … (TRB 4;06) (Clausal relation)

d. The dog girl, but not the same as the other one (THB 3;07) (Other modification)

(Pérez-Leroux, 2012)

But it's hard to know for sure whether the children who make these answers are actually capable of producing second-level recursion on their own -- at least semantically, these answers are most likely to refer to the same object as the target answer that correctly uses the recursive structure of possession, but the children choose their preferred way of describing it -- children really don't have second-level recursion. Or do they, it's just that the recursion is more complicated, so they choose the easier way? It is worth mentioning that this experiment is coded separately from semantic and syntactic angles. The advantage is that apart from correct and wrong, it can find out what other means children have adopted to answer, such as Possessive Pronoun, clausal relation, etc., which makes the error cause analysis relatively perfect.

For the Japanese part, Terunuma (2017), Terunuma &Nakato (2020) and Pérez-leroux (2023) are selected for comparison. It can also be found that the answering rate of the two studies was higher than that of the referential-elicitation task, probably because the answering task only examined whether the child understood the answer, not whether the child could produce the answer on his own, as follows:

2-LOC:

Doobutsuen-no kuruma-no neko-wa nani-iro kana?

zoo-Loc car-Loc cat-Top what-color

Q 'What color is a cat on the car in the zoo?

(Terunuma, 2017)

As you can see, there are fewer tasks and a lighter load on the child's memory, but with more references, the child can also become distracted and miss listening. In addition, isn't it too one-sided to equate acquisition with comprehension? In some cases, a child can understand, but not produce on his or her own? This has not been proven.

In Mandarin related research, truth value judgment and performance task are mainly used. Among them, Mao et al. (2024) and Maria Teresa Guasti et al. (2023) adopted different research methods, and the experimental sample size and age coverage of the two were different, but the final conclusions were the same.

In addition, most of the experimental papers focus on the child's learning path, and the theoretical analysis of the factors behind it is lacking. Taking Roeper and Nakato (2013) as an example, the authors made two conjectures about the child's acquisition path and speculated based on the projection principle combined with the tree diagram, but this has not been directly confirmed by biological evidence. In addition, at present, the tested children are all children with normal intelligence, and there is a lack of research on special children. Based on this, future research should improve experimental methods and combine ERP with advanced experimental means such as EEG or acoustic spectrum analysis to find direct evidence affecting recursive learning mechanism from EEG.

## 7. Conclusion

The purpose of this review is to summarize the experimental paradigms and research contents of tail recursion. It is found that the current tail recursion research mainly focuses on behavioural experiments, with individual prosody and ERP research. The content mainly focuses on the acquisition of locative recursion and possessive recursion in monolingual children of different languages, and a small part of it involves bilingual research. Most studies believe that recursive learning is gradual, and a few scholars believe that it has synchronization. There is no consensus on the cause of the recursion mechanism. Future research should use more advanced technology such as EEG to find more direct biological evidence and combine it with theory to further explore the law of tail recursive acquisition in children.

Figure 1. Levels of tail-recursion sequences

| Language | Studies | Locative | Possessive | 2 level | 3 level | 4 level |
|---|---|---|---|---|---|---|
| English | Pérez-Leroux et al.(2012) | | ○ | ○ | | |
| | Sevcenco, A., Roeper, et al.(2015) | ○ | | ○ | | |
| | Tyler Peterson et al.(2015) | ○ | | ○ | | |
| | Pérez-Leroux et al. (2018) | ○ | | ○ | | |
| | Iain Giblin et al. (2019) | | ○ | ○ | | |
| | Ana T. Pérez-Leroux et al.(2024) | ○ | ○ | ○ | | |
| Japanese | Nakajima et al.(2014) | ○ | ○ | ○ | | |
| | Terunuma et al (2017) | ○ | ○ | ○ | ○ | ○ |
| | Pérez-Leroux et al.(2023) | ○ | ○ | ○ | | |
| Chinese | Shi et al. (2019) | | | ○ | ○ | |
| | Iain Giblin et al. (2019) | | | ○ | ○ | |
| | Li et al. (2023) | | | ○ | ○ | ○ |
| Tamil | Lakshmanan, U. (2022). | ○ | | ○ | | |
| | Usha Lakshmanan (2023) | | ○ | ○ | ○ | ○ |
| Bilingual (Spanish-English) | Pérez-Leroux et al. (2017) | ○ | | ○ | | |
| Bilingual (Hungarian-Romanian) | Larisa Avram et al.(2020) | ○ | | ○ | | |
| French | Roberge et al.(2018) | ○ | ○ | ○ | | |
| German | Pérez-Leroux et al.(2022) | ○ | ○ | ○ | | |
| Hungarian | Toth, A. (2017) | ○ | | ○ | | |
| Karajá | Marcus Maia et al. (2018) | ○ | | ○ | ○ | |
| Kawaiwete | Suzi Lima & Pikuruk Kayabi (2018) | | ○ | ○ | | |
| Persian | Hossein et al.(2023) | ○ | | ○ | ○ | |
| Portuguese | Marcus Maia et al. (2018) | ○ | | ○ | ○ | |
| Piraha | Filomena Sandalo et al. (2018) | ○ | | ○ | ○ | |
| Romanian | Adina Camelia Bleotu (2021) | ○ | | ○ | | |
| Spanish | Pérez-Leroux, A. (2022). | ○ | ○ | ○ | | |

Figure 2. Research purpose, age of participants, and acquisition ratio

| | | | | | | | |
|---|---|---|---|---|---|---|---|
| | 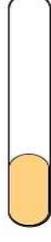 | 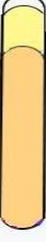 | 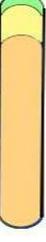 | 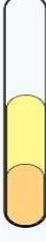 | 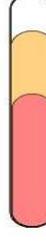 | 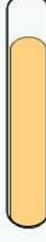 | 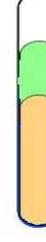 | 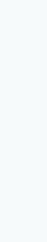 |
| | 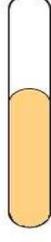 | 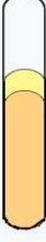 | 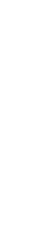 | 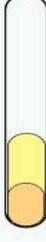 | 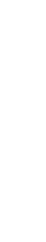 | 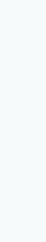 | 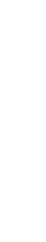 | 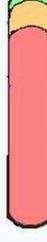 |
| | 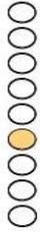 | 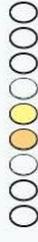 | 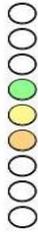 | 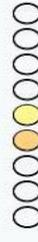 | 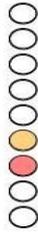 | 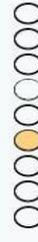 | 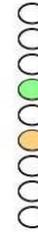 | 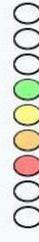 |
| | ○○○○●○○○○ | ○○○●●●○○○○ | ○○○○●●●○○○ | ○○○○●●○○○○ | ○○○●●○○○○○ | ○○○○●●○○○○ | ○○○○●●●○○○ | ○○●●●●○○○○ |
| | Locative recursion & possessive recursion | Developmental path of recursion acquisition Differences of acquisition development of different types | Developmental path of possessive recursion acquisition | Differences in terms of what semantic types of modification relations Children's response pattern compared with adults Preference between *no* structure over relative clauses | Double embedding possessive recursion structure with *de* | The emergence of recursion in children acquiring grammars that incorporate recursion using different syntactic devices | Children's acquisition of possessive recursion in Mandarin Chinese and English | Step-by-step acquisition path of children's DeP recursion structure |
| | Nakajima et al.(2014) | Terunuma et al (2017) | Terunuma & Nakato (2020) | Pérez-Leroux et al.(2023) | Shi et al. (2019) | Iain Giblin et al. (2019) | Li et al. (2023) | Mao et al. (2024) |

Japanese

Chinese

| Language | Author | Topic | | |
|---|---|---|---|---|
| French | Roberge et al.(2018) | Performance and difficulty in recursion acquisition | | |
| German | Pérez-Leroux et al.(2022) | The influence of structural variants on recursion acquisition | | |
| Hungarian | Toth, A. (2017) | Children's explaining of recursion<br>The role of functional head | Pre-school children<br>Grade 2 children | |
| Persian | Hossein et al.(2023) | 3 steps for recursion acquisition | | |
| Romanian | Adina Camelia Bleotu (2021) | Conjunction & recursion | | |
| Spanish | Pérez-Leroux, A. (2022). | Developmental timeline of recursive nominal modification in Spanish | | |

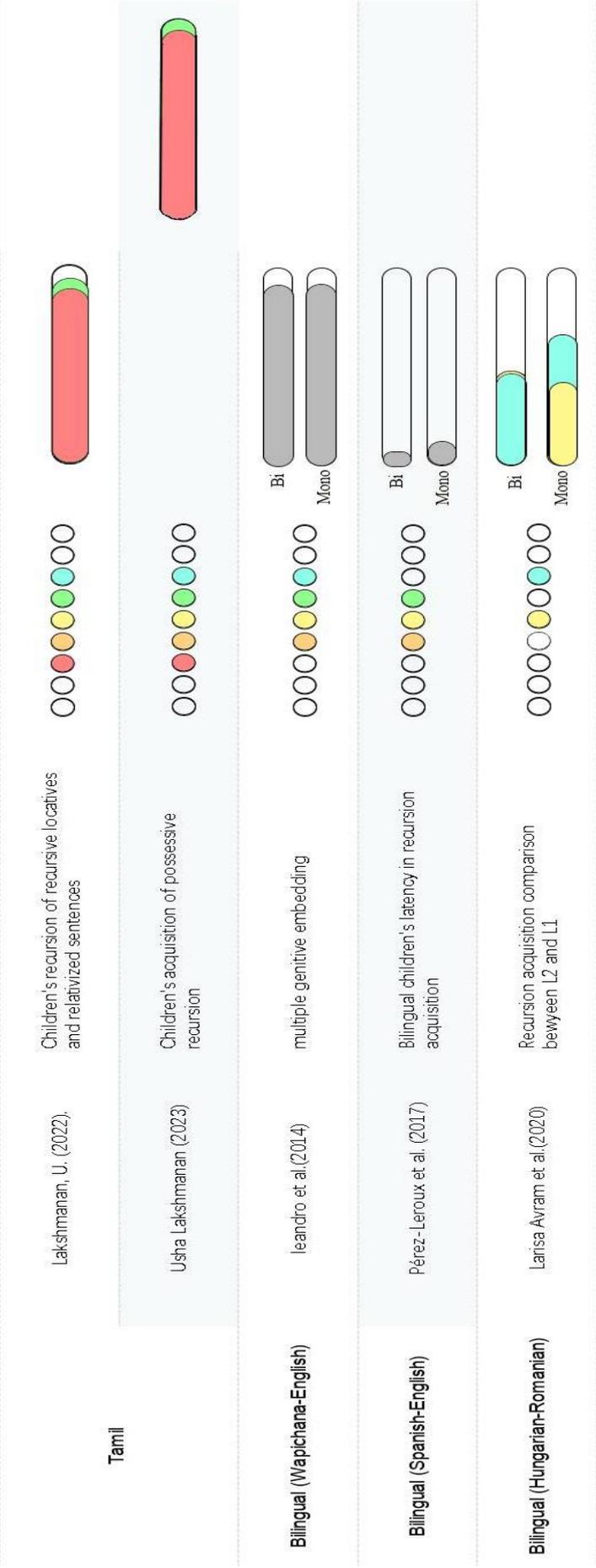

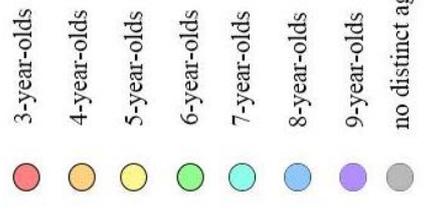